# Robust Perception and Navigation of Autonomous Surface Vehicles in Challenging Environments

Mingi Jeong[1]

The primary objective of my research is to **develop and validate an innovative and robust full autonomy pipeline for a robotic decision-making system** (obstacle detection, tracking, and avoidance) that successfully performs environmental monitoring tasks in **challenging marine environments** (See Fig. 1).

Coastal watersheds globally provide crucial ecological, cultural, and economic benefits, supporting 40% of the world's population and significant biodiversity, fishing, and trade. However, climate change has impacted the region, altering the ecosystem's structure and function. Thus, monitoring and restoring the coastal natural resources that are at risk becomes more critical to cope with the challenges.

Research on coastal regions traditionally involves methods like manual sampling, monitoring buoys, and remote sensing, but these methods face challenges in spatially and temporally diverse regions of interest. Autonomous surface vehicles (ASVs) with artificial intelligence (AI) are being explored, recognized by the International Maritime Organization (IMO) as vital for future ecosystem understanding [27]. However, there is not yet a mature technology for autonomous environmental monitoring due to typically **complex coastal situations**: (1) many static (e.g., buoy, dock) and dynamic (e.g., boats) obstacles not compliant with the rules of the road (COLREGs [7]); (2) uncharted or uncertain information (e.g., non-updated nautical chart); and (3) high-cost ASVs not accessible to the community and citizen science while resulting in technology illiteracy.

To address the above challenges, my research involves both system and algorithmic development: (1) a **robotic boat system** for stable and reliable in-water monitoring, (2) **maritime perception** to detect and track obstacles (such as buoys, and boats), and (3) **navigational decision-making** with multiple-obstacle avoidance and **multi-objective optimization**.

## A. Robotic Boat System and Collaborative Work

Design and build of ASVs have been studied (e.g., [25, 1, 3, 13, 28, 15, 4, 20, 21]) for several years since the first prototype [19]. In literature, the main dimensions for the designs of ASVs are seaworthiness, maneuverability, and control. There is a lack of a systematic analysis of the interaction between *design and sensing performance* for *specific research tasks* of ASVs, e.g., environmental monitoring.

I proposed a novel approach to optimize the design of a research-oriented ASV [12], to modularize the configuration,

[1]Computer Science Department, Dartmouth College, Hanover, NH USA
mingi.jeong.gr@dartmouth.edu

and to enable simple deployment (See Fig. 1 (a)). Following 3D modeling with hydrostatic and computation fluid dynamic tests, we validated the design through field experiments (in lake and sea waters) to measure differences in sensor data according to various designs. While many operate ASVs, our study is the first to analyze the effects of the design on in-water sensors. The proposed method was applied to the custom-built ASV with a water quality sensor to monitor cyanobacterial blooms. Moreover, our qualitative and quantitative analysis methodology is applicable to other ASV design configurations made for any autonomous in-water data collection tasks e.g., coral reef surveying.

Based on the proposed design and approach, I demonstrated utility of a heterogeneous system of robots [24] to advance water quality monitoring in collaboration with biologists. The results show that underwater and ASV sampling, when performed with robust validation and quality assurance/quality control, can lead to the collection of high-quality, spatially and temporally resolved, water quality data.

## B. Efficient Maritime Perception for Situational Awareness

Perception is key to maritime situational awareness and arriving at safe and efficient navigation decisions. As ASVs are mainly deployed in unknown local environments [23], with incomplete information on existing obstacles, approaches that use static information [26] or rely heavily on prior information [5] are not suitable for ASVs. Obstacle detection in the marine domain presents challenges due to the *unstable motion* of both detecting vehicles and detected objects caused by *water dynamics*, as well as the much greater distance and *sparser sensor measurements* (e.g., LiDAR) [14, 6] compared to the ground domain, due to marine traffic safety and vehicle maneuverability. Also, publicly available datasets in the aquatic domain are predominantly single-modal, i.e., image-based, which hinders the development of ASVs compared to other domains, e.g., autonomous cars.

To ensure reliable and fast in-water obstacle detection (fundamental for subsequent tracking and planning), I proposed LiDAR-based efficient in-water obstacle detection [9] and collaborated to develop and test a framework with a Gaussian Mixture probability hypothesis density-based filter tracking [22] (See Fig. 1 (b)). To reduce the computational requirement, our method first converts the 3D point cloud into a 2D spherical projection image. Then, our algorithm based on the integration of a breadth-first search and a variant of an agglomerative clustering segments the points. Our method addresses the sparsity and instability of the point cloud in the aquatic domain – a characteristic that makes the methods developed for self-driving cars not directly applicable for in-water obstacle segmentation. Our method is validated both in simulation and in real-world ASV deployments, with different objects and encountering scenarios. The proposed method is effective in segmenting in-water obstacles not known a priori, in real-time ($36 \pm 16$ ms), outperforming other state-of-the art methods (mIoU – mean Intersection of Union – ours: $84.47\%$; state-of-the-art methods: $18.93\% - 26.25\%$).

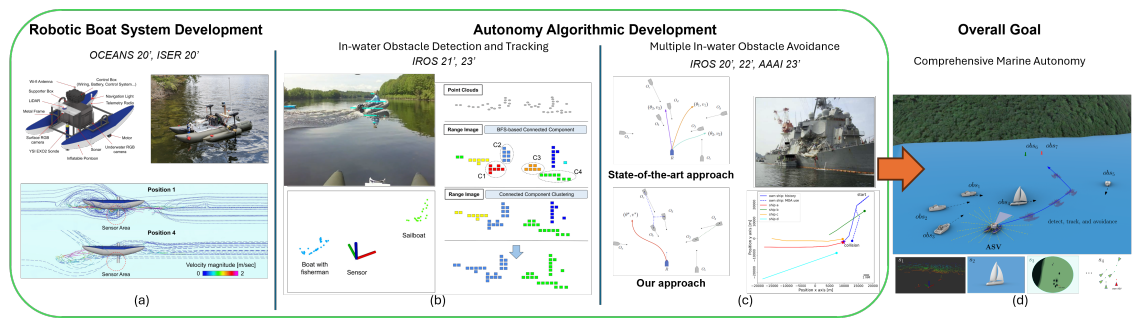

Fig. 1. **(a)** Robotic boat system development for reliable sensing; **(b)** Maritime perception module; **(c)** Multiple obstacle avoidance module in heavy traffic; **(d)** Full pipeline marine autonomy: with the fusion of $n$ onboard sensors (including $s_1$ LiDAR, $s_2$ camera, $s_3$ RADAR,$\cdots$, $s_n$ AIS).

### C. Navigational Decision-making in Heavy Traffic

Navigation and obstacle avoidance in aquatic environments for ASVs in high-traffic scenarios is still an open challenge, as COLREGs is not defined for multi-encounter situations. Current state-of-the-art methods for ASVs operate following COLREGs on several *single-to-single encounters with sequential actions* (e.g., [18, 17]) and *reciprocal cooperative actions by the obstacles* (e.g., [16, 2]). However, these myopic methods may produce conflicting actions ('*give-way*' vs. '*stand-on*') in real-world scenarios [7]. In a real-world navigation, there are many encounters where evasive actions depart from the rule for safety and efficiency, particularly in complex high-traffic scenarios [29].

To adapt to these real-world challenges, first, I have developed a framework based on a novel concept termed *risk vector*, such that collision avoidance decision-making considers a geometric relationship of different encounter situations [8]. Risk vector calculations based on ego-centric perspective allow the proposed method to perform in real-time. For example, in a single obstacle scenario, our algorithm performs $1.68$ times faster computation as well as more efficient action resulting in similar clearance $14°$ heading alteration at the same speed compared while $26°$ heading alteration with $40\%$ speed reduction compared to the other methods.

By extending to multiple scenarios (See Fig. 1 (c)), my work [10] proposes a novel real-time non-myopic obstacle avoidance method, allowing an ASV to achieve a holistic view of the feasible action space and avoid deadlock scenarios common in conventional approaches [17, 18]. My approach proposes (1) a *clustering method* based on motion attributes of other obstacles, (2) a *geometric framework* and *theoretical proof* to identify a feasible action space, and (3) a *multi-objective optimization* to determine the best action derived with *Pareto-optimality*. The proposed method shows $1.2$–$1.7$ times higher success rate performance ($0.927$ on average) than other methods, while achieving real-time computation ($61 \pm 13$ ms) in congested traffic with 30 obstacles. We validated the proposed approach by (1) extensive Monte Carlo simulations (total 2.000 runs); (2) historical Automatic Identification System (AIS) data for a marine accident; and (3) real in-water experiments by our custom ASV. The overall system demonstration was also published and received many attractions at a top-tier AI conference [11].

### D. Future Work

Based on extensive research outcomes, I am committed to advancing and validating comprehensive marine autonomy in field robotics, which promises significant societal benefits (Figure 1(d)). The results of my research, including data, will be openly shared, in line with my prior tracks, to foster equity, inclusion, and accessibility.

*1) Low-cost robot for community:* I aim to create a fleet of affordable robotic boats, outfitted with advanced sensors, to monitor aquatic environments efficiently. These boats will measure various physical, biological, and chemical parameters, identified in collaboration with stakeholders and experts. This initiative addresses key challenges in the scientific community: (1) the high cost of the platform (typically in the order of $> 10$k USD), (2) non-user friendly interface, and (3) missing automated data processing pipelines.

*2) Enhanced In-water Obstacle Detection through Sensor Fusion:* I propose to develop a novel sensor fusion algorithm that adaptively combines data from multiple sensors (such as cameras and affordable range sensors). The hierarchical architecture with prioritization will enable precise tracking of both stationary and moving objects, improving our understanding of their movements. Additionally, I will release the first maritime dataset featuring multi-modal sensor dataset, capturing a wide range of in-water objects and environmental conditions.

*3) Collision Avoidance by Topological Intent-Awareness:* I will further develop an adaptive COLREGs-compliant method to robustly address uncertain encounter scenarios. By employing a learning-based strategy that utilizes the concept of homotopy, this method will improve the ASV's ability to interpret the intentions of nearby obstacles, leveraging real-time and historical data. This innovative approach aims to significantly reduce uncertainty in obstacle intent, a crucial consideration in maritime navigation.